\documentclass{bmvc2k}
\usepackage{amssymb}
\usepackage{lipsum}
\usepackage[table]{xcolor}
\usepackage{algorithm}
\usepackage{enumitem}

\usepackage[noend]{algpseudocode}
\usepackage{threeparttable}

\title{SIA-GCN: A Spatial Information Aware Graph Neural Network with 2D Convolutions for Hand Pose Estimation}

\addauthor{Deying Kong}{deyingk@uci.edu}{1}
\addauthor{Haoyu Ma}{haoyum3@uci.edu}{1}
\addauthor{Xiaohui Xie}{xhx@ics.uci.edu}{1}

\addinstitution{
 University of California, Irvine\\
 USA
}

\runninghead{Student, Prof, Collaborator}{BMVC Author Guidelines}

\begin{document}

\maketitle
\begin{abstract}
Graph Neural Networks (GNNs) generalize neural networks from applications on regular structures to applications on arbitrary graphs, and have shown success in many application domains such as computer vision, social networks and chemistry. In this paper, we extend GNNs along two directions: a) allowing features at each node to be represented by 2D spatial confidence maps instead of 1D vectors; and b) proposing an efficient operation to integrate information from neighboring nodes through 2D convolutions with different learnable kernels at each edge. 
The proposed SIA-GCN can efficiently extract spatial information from 2D maps at each node and propagate them through graph convolution. By associating each edge with a designated convolution kernel, the SIA-GCN could capture different spatial relationships for different pairs of neighboring nodes.
We demonstrate the utility of SIA-GCN on the task of estimating hand keypoints from single-frame images, where the nodes represent the 2D coordinate heatmaps of  keypoints and the edges denote the kinetic relationships between keypoints. 
Experiments on multiple datasets show that  SIA-GCN provides a flexible and yet powerful framework to account for structural constraints between keypoints, and can achieve state-of-the-art performance on the task of hand pose estimation.

\end{abstract}

\section{Introduction}



Hand pose estimation is a long standing research area in computer vision, given its vast potential applications in computer interaction, augmented reality, virtual reality and so on~\cite{DoostiSurvey}.
It aims to infer 2D or 3D positions of hand keypoints from a single input image or a sequence of images, which could possibly take the form of RGB, RGB-D or grayscale. 
Although 3D hand pose estimation is drawing increasing attention in the research community~\cite{wang2019geometric, malik2020handvoxnet, xiong2019a2j,wan2019self,yuan2018depth,ge2018point}, 2D hand pose estimation still remains a valuable and challenging problem~\cite{simon2017hand,wang2018mask,kong2019adaptive}.
A plentiful of 3D hand pose estimation algorithms rely on their 2D counterparts~\cite{cai2018weakly, zimmermann2017learning}, attempting to lift 2D predictions to 3D space.
In this paper, we investigate the problem of 2D handpose estimation from single RGB image.

The progress in hand pose estimation research has been boosted greatly by the  invention of deep Convolutional Neural Networks (CNNs). Deep CNN models like Convolutional Pose Machine~\cite{wei2016convolutional} and Stacked Hourglass~\cite{newell2016stacked} have been successfully applied to 2D hand pose estimation, though they are originally proposed to solve the task of human pose estimation. 
Some methods~\cite{kong2020rotation, kong2019adaptive, chen2014articulated} also integrate deep CNNs with probabilistic graphical model to harvest both the powerful representation ability of deep CNNs and the capability of explicitly expressing spatial relationships attributed to graphical model.

In contrast to CNN, graph neural network has the ability to handle irregular structured data.
The joints of a human body, and keypoints of a hand can be conveniently considered as irregular graphs, giving possibilities of applying Graph Convolutional Network (GCN)~\cite{kipf2016semi} on human/hand pose estimation tasks. 
However, in the vanilla GCN~\cite{kipf2016semi}, all the nodes share the same one-hop propagation weight matrix, which makes it unready to be applied to pose estimation task because different human body joints and bones should have different semantics.
Authors in~\cite{doosti2020hope, zhao2019semantic, cai2019exploiting} have proposed different variants of the vanilla GCN from~\cite{kipf2016semi} for the purpose of human or hand pose estimation. 
However, all these methods take as input a one dimensional vector for each node, and the node feature at each layer is always a one dimensional vector. Thus, they are not ready to process 2D confidence map. Although, in~\cite{doosti2020hope, zhao2019semantic, cai2019exploiting}, modifications are made to vanilla GCN, they still do not allow full independence among the edges.

In this paper we propose the Spatial Information Aware Graph Neural Network with 2D convolutions (SIA-GCN). In SIA-GCN, the feature of each node is a two dimensional matrix, and the information propagation to neighboring nodes are carried out via 2D convolutions along each edge. By using 2D convolutions instead of flattening the 2D feature map to a 1D vector and then performing linear multiplications, the spatial information encoded in the feature map is reserved and appropriately exploited. We also propose to use different 2D convolutional kernels on different edges, aiming to capture different spatial relationships for different pairs of neighboring nodes. 
The SIA-GCN is very flexible and could be easily combined with off-the-shelf 2D pose estimators. 
In this work, we demonstrate the efficacy of SIA-GCN on 2D hand pose estimation.
For this application, the 2D feature maps at the nodes are actually the confidence maps of the hand keypoint positions.  With a designated matrix for each edge, the SIA-GCN has the ability to capture various spatial relationships between different pairs of hand keypoints.


Our main contributions are threefold: 
\begin{itemize}[noitemsep, topsep=1pt]
    \item We propose the novel SIA-GCN which can process 2D confidence maps for each node efficiently and effectively, by integrating graph neural networks and 2D convolutions. 
Using 2D convolutions, our SIA-GCN can exploit and harvest the spatial information provided in the 2D feature maps.
\item By assigning different convolutional kernels on different edges, the SIA-GCN has the property of full edge-awareness. Distinct spatial relationships can be learned on different edges. 
\item We deploy SIA-GCN in the task of hand pose estimation. Utilizing SIA-GCN, the constructed neural network can achieve state-of-the-art performance.
\end{itemize}



\section{Related Work}
There exists a vast amount of research focusing on topics of human/hand pose estimation~\cite{simon2017hand, wang2019geometric,malik2020handvoxnet, xiong2019a2j, wan2019self, wang2020predicting, yuan2018depth, baek2018augmented, wan2018dense, ge2018point, mueller2017real,ge2016robust} and graph neural networks~\cite{simon2017hand, doosti2020hope, zhao2019semantic, cai2019exploiting}. In the related work, we focus on 2D hand pose estimation from single RGB images and graph convolutional network~\cite{simon2017hand}'s applications to pose estimation tasks.

\vspace{0.5em}
\noindent\textbf{2D hand pose estimation.} Studies of RGB image based 2D hand pose estimation has long benefited from that of human pose estimation, where deep Convolutional Neural Networks (CNNs) have enjoyed great success ~\cite{toshev2014deeppose, wei2016convolutional, newell2016stacked, xiao2018simple, chen2018cascaded, sun2019deep}. Among these deep CNN models, Convolutional Pose Machines~\cite{wei2016convolutional} and Stacked Hourglass~\cite{newell2016stacked} are commonly used in various RGB-based 2D hand pose estimation methods~\cite{simon2017hand, kong2020rotation, kong2019adaptive, chen2020nonparametric, wang2018mask}.
Compared with deep CNNs, Graphical Model (GM) has also played a significant role in solving the pose estimation task. GM has the power of modeling spatial constraints among the joints explicitly.
Recently, several works in pose estimation combine GM and neural network to fully exploit the structural information~\cite{tompson2014joint, chen2014articulated,song2017thin, yang2016end, kong2019adaptive, kong2020rotation}. 
Traditionally, GM with fixed parameters~\cite{tompson2014joint,song2017thin,chen2014articulated} are applied to the pose estimation task, while 
the most recent work in~\cite{kong2019adaptive, kong2020rotation} propose to adopt GM with adaptive parameters conditioning on input images. Although all take advantage of structural information, our proposed method is based on graph convolutional network while these previous works~\cite{kong2019adaptive, kong2020rotation} are based on graphical models. 

\vspace{0.5em}
\noindent\textbf{Graph convolutional network.}
Graph Convolutional Network (GCN), which generalizes deep CNNs to graph structured data, have attracted increasing attention in recent years. One main research direction is to define graph convolutions from the spectral perspective~\cite{shuman2013emerging}, while the other works on the spatial domain~\cite{kipf2016semi}.  For a comprehensive survey on GCN, we refer readers to~\cite{wu2020comprehensive}. The most related works to ours are~\cite{doosti2020hope, zhao2019semantic, cai2019exploiting}, in which variants of spatial GCNs have been proposed and applied to human/hand pose estimation tasks in the computer vision field. In the following, we discuss the key differences between our SIA-GCN and those in~\cite{doosti2020hope, zhao2019semantic, cai2019exploiting}.

In~\cite{cai2019exploiting}, the authors have proposed to classify neighboring nodes according to their semantic meanings and use different kernels for different neighboring nodes. The purpose of their proposed GCN is to regress 3D position vectors from 2D position vectors, and the input to the GCN for each node is a one dimensional $\mathbb{R}^2$ vector, representing predicted 2D position of a corresponding body joint. However, our proposed SIA-GCN aims to handle two dimensional confidence maps for each node. The confidence map inherently contains much more information than the two-element position vector. Our goal is to refine final 2D predictions, other than lifting 2D predictions to 3D space. Besides, instead of classifying nodes into different classes, we treat every edge independently and attach a designate weight kernel to each edge.

In~\cite{doosti2020hope}, the authors directly adopt the propagation rule from~\cite{kipf2016semi} with the modification that, instead of using a predefined adjacency matrix, they have proposed to use an adaptive adjacency matrix which could be learned from data. The feature for each node is a one dimensional vector. Our method differs from~\cite{doosti2020hope} in that edge-dependent weights are considered explicitly and our SIA-GCN works on 2D confidence maps for each node.

In~\cite{zhao2019semantic}, the proposed Semantic Graph Convolution (SemGConv) adds a learnable weighting matrix to conventional graph convolutions from~\cite{kipf2016semi}. The weight matrix serves as a weighting mask on the edges of a node when information aggregation is performed. The SemGConv is inherited from ST-GCN~\cite{yan2018spatial}, but is equipped with additional important features such as softmax non-linearity and channel wise masks. The weighting mask adds a scalar importance weight (or a vector if it's channel wise) to each edge. However, in SIA-GCN, we directly attach to each edge a fully independent convolution matrix. Besides, our SIA-GCN works on 2D node features with spatial information awareness.

\section{Methodology}
\begin{figure}[h]
    \centering
    \includegraphics[width=0.85\textwidth]{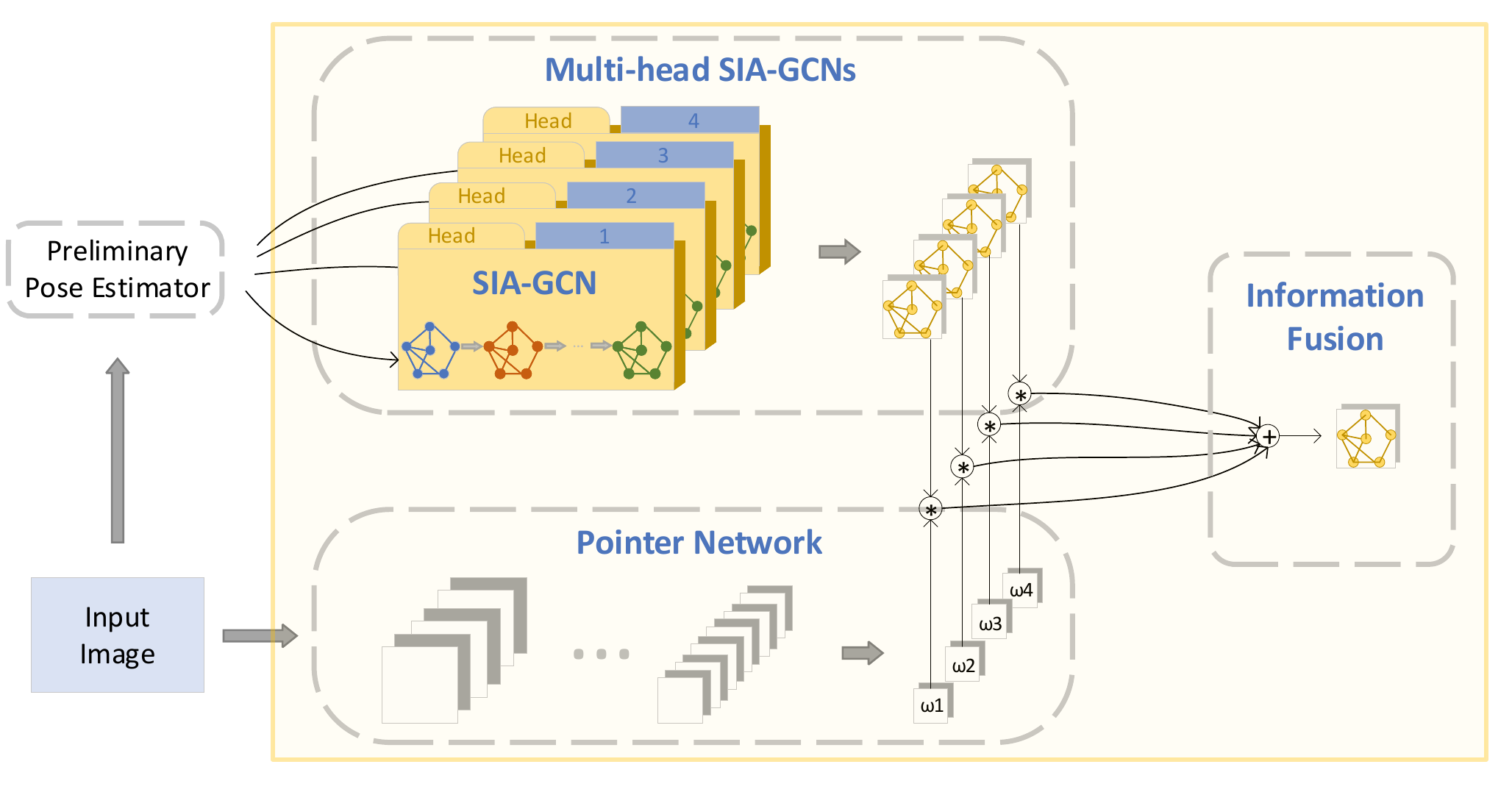}
    \caption{System diagram of the SiaPose, utilizing SIA-GCN.}
    \label{fig:system_diagram}
\end{figure}

In this section, we present the SIA-GCN, and its application to hand pose estimation. We refer to the resulted pose estimator as SiaPose, which is illustrated in Fig~\ref{fig:system_diagram}.

The SiaPose takes as input a RGB image, to which a preliminary pose estimator is applied. The preliminary pose estimator could be any 2D pose estimator, such as the famous Convolutional Pose Machine~\cite{wei2016convolutional} and Stacked Hourglass~\cite{newell2016stacked}, which would output a set of confidence maps of keypoint positions. 
Then, at the top branch, the confidence maps are fed into a block of multi-head SIA-GCNs. 
Each SIA-GCN processes a copy of the confidence maps parallelly and independently.
Meanwhile  at the bottom branch, the input image goes through a pointer network, which gives a weight vector, indicating which head is important in the multi-head SIA-GCNs.
Finally, at the information fusion stage, confidence maps output from the multi-head SIA-GCNs are aggregated according to the weight vector. 



In the following subsections, we revisit the graph convolutional network first, and discuss the motivation for our SIA-GCN. Then, we present a compact formulation of our proposed edge-aware graph convolutional layers in SIA-GCN, and  demonstrate how to implement it efficiently using 2D convolutional operations.
Finally, we describe the training procedure of the SiaPose.

\subsection{Revisiting Graph Convolutional Network}
\label{subsection:revisit_gcn}
The Graph Convolutional Network (GCN) proposed in~\cite{kipf2016semi} has enjoyed great success on a variety of  applications since its advent. Given a graph $\mathcal{G}=(\mathcal{V}, \mathcal{E})$ with $N$ nodes $v_i \in \mathcal{V}$, edges $(v_i, v_j) \in \mathcal{E}$, adjacency matrix $A \in \mathbb{R}^{N \times N}$, and a degree matrix $D \in \mathbb{R}^{N \times N}$ with $D_{ii} = \sum_{j}A_{ij}$, 
the layer-wise propagation rule is characterized by the following equation

\begin{equation}
    H^{(l+1)} = \sigma \left(\Tilde{D}^{-\frac{1}{2}} \Tilde{A}\Tilde{D}^{-\frac{1}{2}}H^{(l)}W^{(l)}\right),
\label{eq:1}
\end{equation}
where $\Tilde{A} = A + I_N$ is the adjacency matrix of the undirected graph $\mathcal{G}$ with self-connections~\cite{kipf2016semi}.
$I_N$ is the identity matrix, $\Tilde{D}_{ii} = \sum_{j}\Tilde{A}_{ij}$. $H^{(l)} \in \mathbb{R}^{N \times M}$ is the matrix of activations in the $l^{th}$ layer, or input feature matrix of the $l^{th}$ layer. The parameter $W^{(l)}$ is the trainable weight matrix of layer $l$.

In the scenario of human and hand pose estimation, it is well studied that probabilistic graphical models could be deployed to enhance structural consistency~\cite{tompson2014joint,kong2020rotation,chen2014articulated}. The graphical model could take in some preliminarily generated 2D confidence maps of each body joint or hand points. These confidence maps are usually considered as the unary potential functions by the graphical model. Then the graphical model could impose some learned pairwise potential functions on the initial confidence maps, thus enforcing spatial consistency of the body joints/keypoints. Can we also apply GCN to the confidence maps and then enhance spatial consistency?

The answer is positive, but it's not trivial. To apply the above GCN to pose estimation, some modifications are needed due to the dimensionality. In Eq.~(\ref{eq:1}), the activation matrix $H^{(l)} \in \mathbb{R}^{N \times M}$  is a two dimensional matrix, corresponding to $N$ nodes and each node is associated with a 1-d feature of size $M$. 
However, for the case of 2D pose estimation, each graph node (usually corresponding to a joint or keypoint) can be associated with a two dimensional confidence map. 
This discrepancy could be handled by flattening the two dimensional confidence map to a single long vector and then perform layer propagation according to Eq.~(\ref{eq:1}). 
However, this would result in very large feature size, significantly increase the computational complexity (imagine that a $64 \times 64$ matrix would result in a one dimensional vector of size 4069).
Besides, by flattening the confidence map, spatial information encoded in the confidence map would be corrupted.
Thus, we propose to use 2D convolutional operations directly on 2D confidence maps when propagating information along the edges.

Moreover, in Eq.~(\ref{eq:1}), since all the node share the same weight matrix $W^{(l)}$ and information aggregation is only controlled by the adjacency relationships between nodes, it would be difficult for the propagation rule in Eq.~(\ref{eq:1}) to characterize different positional relationships for different pairs of neighboring joints. For example, the positional information propagation between two neighboring thumb joints should be different from that between the neighboring joints on the middle finger. One simple reason is that the bones from the thumb and middle finger actually have different lengths.  

\subsection{SIA-GCN}
\begin{figure}[!h]
    \centering
    \includegraphics[width=0.42\textwidth]{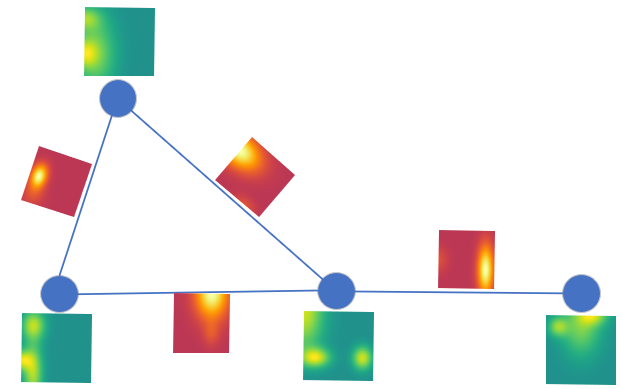}
    \caption{A simple illustration of SIA-GCN.}
    \label{fig:SIA-GCN}
\end{figure}

To resolve the above mentioned concerns, we propose the spatial information aware graph neural network with 2D convolutions (SIA-GCN), where each edge of the graph is associated with an individual learnable 2D convolutional kernel. A toy example of a graph consisting of four nodes is shown in Fig.~\ref{fig:SIA-GCN}, where green matrices represent 2D features (heatmaps) at each node and red matrices represent designated 2D kernels associated with each edge.  

For the task of hand pose estimation, we could define a graph $\mathcal{G} = (\mathcal{V}, \mathcal{E})$ where $\mathcal{V} = \{v_1, v_2, \cdots v_K\}$ is the set of nodes corresponding to $K$ hand keypoints, and $\mathcal{E}$ is the set of edges encoding the neighboring relationships among the keypoints. Each node $v_i$ is associated with a 2D confidence map $X_i \in \mathbb{R}^{ h \times w}$, which encodes the positional information of $i^{th}$ keypoint. We could stack all $\{X_i \}$ for $i=1,2,..,K$ in a 3D matrix, and denote it as $X \in \mathbb{R}^{K \times h \times w}$. 

One important feature of our SIA-GCN is that each edge in $\mathcal{E}$ is associated with an individual weight matrix or 2D convolutional kernal, $F_j \in \mathbb{R}^{h' \times w'}$, $j = 1,2,\cdots, |\mathcal{E}|$. 
Again, we compact all $\{F_j\}$ into a single matrix $F\in \mathbb{R}^{|\mathcal{E}| \times h' \times w'}$, which is actually the set of learnable parameters of the edge-aware graph convolutional layer.
The information propagated from node $i$ to node $j$ along edge $e_{i,j}$ is obtained by calculating the 2D convolution of $X_i \circledast F_{{e}_{i,j}}$. 
Then, all the information propagated into node $i$ are aggregated according to the adjacency matrix.
The propagation rule could be presented compactly in matrix multiplications and convolutions as
\begin{equation}
    X^{(l+1)} = \sigma \left( \hat{A} \left((BX^{(l)}) \tilde{\circledast} {F}^{(l)}\right)\right),
    \label{eq:our_prop}
\end{equation}
where the superscript $l$ and $l+1$ denote the $l^{th}$ layer and $l+1^{th}$ layer respectively, $\tilde{\circledast}$ is the channel-wise 2D convolution operator, and $\sigma(\cdot)$ is the non-linear activation function. The matrix $B \in \mathbb{R}^{|\mathcal{E}| \times K}$ is the broadcast matrix, which broadcasts node features to its outgoing edges. Note that the matrix multiplication $B X^{(l)}$ results in a shape of $|\mathcal{E}| \times h \times w$, whereas originally the dimension of $X^{(l)}$ is $K \times h \times w$. 
In other words, the operation $B X^{(l)}$ simply prepares the input along each edge for the following channel-wise convolution, $(BX^{(l)}) \tilde{\circledast} {F}^{(l)}$. Finally, the matrix $\hat{A} \in \mathbb{R}^{K \times |\mathcal{E}|}$ is the aggregation matrix, which harvests all the information from the incoming edges to the graph nodes. 

It is worth pointing out that, in Eq.~(\ref{eq:our_prop}), only $F^{(l)}$ is the learnable parameter, while the broadcast matrix $B$ and the aggregation matrix $\hat{A}$ are both determined and constructed from the graph's adjacency matrix $A$ by  Algorithm~\ref{alg: broadcastMatrix}. In Algorithm~\ref{alg: broadcastMatrix}, we assume the input adjacency matrix $A$ is already included with self connections.

\begin{figure}[ht]
  \centering
  \vspace{-1em}
\begin{minipage}{.9\linewidth}
\begin{algorithm}[H]

\caption{Broadcast and Aggregation Matrices Construction}\label{alg: broadcastMatrix}
\begin{algorithmic}[1]

\Procedure{ConstructMatrices}{$A$}
\Comment{Input $A$ is the adjacency matrix}
\vspace{0.1em}
\State Find the number of directed edges, $|\mathcal{E}|$, from $A$
\State Find the number of nodes, $K$, from $A$
\vspace{0.1em}
\State \textbf{Initialize} \Comment{Initialization for $B$ and $\hat{A}$}
\State \hspace{5 mm}$B$ as a zero matrix of size $|\mathcal{E}| \times K$
\State \hspace{5 mm}$\hat{A}$ as a zero matrix of size $K \times |\mathcal{E}|$
\State \hspace{5 mm}${e}$ as a zero vector of size $|\mathcal{E}|$
\State \hspace{5 mm}$m=1$ 

\vspace{0.1em}
\For{$i$ in $1,2,\cdots,K$} \Comment{Calculate for $B$}
\For{$j$ in $1,2,\cdots,K$}
\If{$A_{j, i}==1$}\Comment{If $j$ is the starting node of edge $m$}\\
\hspace{20 mm} $B_{m,j}=1$ \\
\hspace{20 mm} $e[m]=i$ \Comment{Record the end node of edge $m$}\\
\hspace{20 mm} $m=m+1$
\EndIf
\EndFor
\EndFor

\vspace{0.1em}
\For{$m$ in $1,2,\cdots,|\mathcal{E}|$} \Comment{Calculate for $\hat{A}$}

\State $\hat{A}_{e[m],m}=1$  

\EndFor

\vspace{0.1em}
\State Construct the diagonal degree matrix $D$, with $D_{ii} = \sum_j \hat A_{ij}$.
\State Set $\hat{A}= D^{-1}\hat{A}$  \Comment{Normalize $\hat{A}$}

\vspace{0.1em}
\State \textbf{return} $B$, $\hat{A}$
\EndProcedure
\end{algorithmic}
\end{algorithm}
\end{minipage}
\end{figure}







\vspace{-1em}
\subsection{SiaPose and its training procedure}
With SIA-GCN, we propose the SiaPose for 2D hand pose estimation, as in Fig.~\ref{fig:system_diagram}.
The preliminary pose estimator could be any off-the-shelf 2D hand pose estimator. 
Multiple heads of SIA-GCN would benefit capturing different positional informations due to different hand shapes in the input images. Assume there are $M$ heads in the multi-head SIA-GCNs, then, we could denote the output of the multi-head SIA-GCNs as $Y\in \mathbb{R}^{M \times K \times h \times w}$ and the output at the $m^{th}$ SIA-GCN as  $Y_m\in \mathbb{R}^{K \times h \times w}$. The pointer network, whose input is the image, is a regression network which generate a soft pointer vector $w \in \mathbb{R}^{M}$. 
The weight vector $w$ actually indicates the importance of the information generated at different heads. Finally, at the information fusion stage, the aggregated confidence map is given by
\begin{equation}
    \bar Y = w \cdot Y = \sum_{m=1}^{M} w_mY_m,
\label{eq: aggregation}
\end{equation}
which is a weighted sum of ${Y}_m$. The final predictions of the keypoint positions are obtained by taking the argmax of $\bar Y$.

The training procedure of the SiaPose is simple and could be conducted in an end-to-end fashion. 
The total loss function is defined as 
\begin{equation}
    L = \alpha L_1 + L_2 = \alpha \sum_{t=1}^{T} \sum_{k=1}^{K} \left\lVert S_k^{t} - Y_k^{*}\right\rVert _F^2 + \sum_{k=1}^{K} \left\lVert \bar{Y}_k - Y_k^{*}\right\rVert _F^2.
\label{eq:loss_function}
\end{equation}
The first loss $L_1$ is responsible for the output of the preliminary pose estimator, while the second loss $L_2$ is added at the final output. 
The preliminary pose estimator itself (e.g. CPM and Stacked Hourglass) might consist of $T$ multiple stages.
The term $S_k^{t} \in \mathbb{R}^{h \times w}$ is the confidence map of $k^{th}$ keypoint generated by the $t^{th}$ stage of the preliminary pose estimator, while $\bar{Y}$ is the final confidence output of the SiaPose as in Eq.(\ref{eq: aggregation}).
Besides, $Y_k^{*} \in \mathbb{R}^{h \times w}$ is the ground truth confidence map of $k^{th}$ keypoint, created by placing a Gaussian peak at its ground truth position. The coefficient $\alpha$ serves as a balancing weight between the two loss functions.

\section{Experiments}
\textbf{Datasets.}
We evaluate our proposed method on three public hand pose datasets, the CMU Panoptic Hand Dataset (Panoptic) \cite{simon2017hand}, the MPII+NZSL Hand Dataset~\cite{simon2017hand} and the Large-scale Multiview 3D Hand Pose Dataset (MHP) \cite{Francisco2017}. 
For Panoptic (\textasciitilde 15k images) and MHP (\textasciitilde 82k images), we follow the setting of \cite{kong2020rotation} and randomly split all samples into training set (70\%), validation set (15\%) and test set (15\%).
Since our contribution mainly focus on pose estimation instead of detection, we crop square image patches of annotated hands off the original images.
A square bounding box which is 2.2 times the size of
the hand is applied for cropping as in \cite{simon2017hand,kong2020rotation, kong2019adaptive}.

\textbf{Evaluation metrics.} The Probability of Correct Keypoint (PCK)~\cite{simon2017hand} is utilized as our evaluation metric. 
In this paper, we use normalized threshold with respect to the size of square bounding box. We report the performance under different thresholds, $\delta$ = \{0.01, 0.02, 0.03, 0.04, 0.05, 0.06\}, and also their average (mPCK). More formally, for a single cropped input image of size $s\times s$, the PCK at $\delta$ can be defined as 
\begin{equation}
 \textrm{PCK}(\delta) = {N(\delta)}/{K},   
\end{equation}
where $N(\delta)$ is the number of predicted keypoints which are  within an interval threshold $\delta \cdot s$ of its correct location and $K$ is the total number of keypoints.

\textbf{Implementation details.} In the experiments, two baselines, i.e., six-staged Convolutional Pose Machine (CPM) as in~\cite{simon2017hand} and eight-staged Stacked Hourglass (SHG) are used as  preliminary pose estimators in our SiaPose. For the SIA-GCN, we use 5 edge-aware graph convolutional layers defined in Eq.~(\ref{eq:our_prop}), which adopts a tree structured graph according to the kinematic structure of the hand skeleton, adding self connections. The size of the convolutional kernels in Eq.~(\ref{eq:our_prop}) is set to 45.
ResNet-18 is used as the backbone of the pointer network.
The input image is resized to $368 \times 368$ and $256 \times 256$ for the cases of CPM and SHG, respectively.
Images are then scaled to [0,1], and normalized with mean of (0.485, 0.456,
0.406) and standard deviation of (0.229, 0.224, 0.225).
We use Adam as our optimizer. For SHG-based SiaPose, the initial learning rate is set to 7.5e-4 while for the CPM-based SiaPose, we set it to 1e-4.  For both cases, we train the model for 100 epochs, with learning rate reduced by a factor of 0.5 at milestones of the 60-th and 80-th epoch. The weight coefficient $\alpha$ in loss function Eq.~(\ref{eq:loss_function}) is set to drop from 1.0 to 0.1 at the 40th epoch.

\textbf{Comparison with baselines.}
In Table~\ref{tab:SHG_Panoptic} and Table~\ref{tab:CPM_Panoptic}, we compare the performance of our SiaPose with two baselines, CPM and SHG. 
(1) First, we conduct an experiment where \emph{edge-unaware} GCN is utilized, where a shared weight matrix is used for all the edges. Interestingly, it performs worse than the baseline models. This is reasonable, because it's not appropriate to assume that  relative positions of neighboring keypoints are always the same. For example, index finger and thumb naturally have bones with different lengths. 
(2) Then we conduct experiments with our \emph{edge-aware} SIA-GCNs, where different numbers of heads are explored. 
The results demonstrate that our proposed SiaPose could consistently improve both baselines noticeably. The ablative study on different numbers of heads validates the benefit of multi-heads and the effectiveness of the proposed SIA-GCN. For SHG, there is a 2.12 percent improvement at threshold $\delta = 0.01$ and for CPM, a 1.95 percent improvement is seen at threshold $\delta = 0.04$.
(3) Also, inspired by the state-of-the-art algorithm~\cite{kong2020rotation}, by adding a rotation network into our SiaPose (R-SiaPose) and using a similar training strategy, the performance of our method is further boosted, leading to significant improvements from baselines. Improvements of about 5 percent for SHG and nearly 4 percent for CPM are observed. We would also compare our model with that proposed in~\cite{kong2020rotation} in next subsection.

\begin{table}[h]
\small 
    \centering
    \caption{SHG based SiaPose on Panoptic Dataset.}
    \begin{tabular}{c| r| r |r|r|r|r|r}
    \hline
    PCK@ & {0.01}& 0.02& 0.03& 0.04& 0.05& 0.06& mPCK \\
    \hline
    \hline
         SHG Baseline & 35.85 & 71.47 & 83.15 & 88.21 & 91.10 & 92.92 & 77.12 \\
         \hline
         \hline
         SharedWeight GCN  & 34.76 & 69.66 & 81.33& 86.19& 89.14& 90.95& 75.34 \\  
         \hline
         \hline
         1-head SiaPose & {35.78} & 71.16& 83.57& 88.98& 92.00& 93.84& 77.55 \\
        \hline
         5-head SiaPose  & {37.53} & 73.07 & 84.60& 89.51& 92.14& 93.85& 78.45 \\
        \hline
        10-head SiaPose  & 37.97 & 73.53 & 84.95& 89.70& 92.26& 93.91& 78.72 \\ 
        \hline
        Improvement &2.12& 2.06& 1.80& 1.49& 1.16&0.99& 1.60 \\
         \hline
         \hline
        10-head R-SiaPose  & 39.46 & 77.22& 88.45& 92.97& 94.85& 96.09 & 81.48 \\        
         \hline
        Improvement & 3.61 & 5.75 & 5.30 & 4.76 & 3.75 & 3.17 & 4.36\\
        \hline

    \end{tabular}
    \label{tab:SHG_Panoptic}
\end{table}{}

\begin{table}[h]
\small 
    \centering
    \caption{CPM based SiaPose on Panoptic Dataset.}
    \begin{tabular}{c|r|r|r|r|r|r|r}
    \hline
    PCK@ & 0.01& 0.02& 0.03& 0.04& 0.05& 0.06& mPCK \\
    \hline
    \hline
         CPM Baseline & 25.73 & 62.77 & 77.80 & 84.35 & 88.11 & 90.57 & 71.55 \\
         \hline
         \hline
         SharedWeight GCN  & 25.14 & 61.76 & 77.13 & 83.60 & 86.97 & 89.20 & 70.63 \\
         \hline
         \hline
         1-head SiaPose & 25.90 & 63.36 & 78.98& 85.69& 89.44& 91.90& 72.55 \\
        \hline
         5-head SiaPose  & 26.36 & 64.05 & 79.11 & 85.74 & 89.38& 91.78 & 72.74 \\
        \hline
         10-head SiaPose  & 26.45 & 64.19 & 79.67 & 86.30 & 89.83 & 92.20 & 73.11 \\        
         \hline
         Improvement & 0.72& 1.42& 1.87& 1.95& 1.72& 1.63& 1.56\\
         \hline
         \hline
         10-head R-SiaPose & 26.62 & 65.80& 81.60& 88.02& 91.39& 93.36&  74.47\\
         \hline
         Improvement &0.89  & 3.03 & 3.80 & 3.67  & 3.28  & 2.79  & 2.92\\
         \hline 
    \end{tabular}
    \label{tab:CPM_Panoptic}
\end{table}{}

\textbf{Comparison with state-of-the-art methods.}
We further compare our approach with the current state-of-the-art methods~\cite{kong2020rotation, kong2019adaptive}. Probabilistic graphical models are deployed in~\cite{kong2020rotation} and~\cite{kong2019adaptive}, where the output confidence maps from CPM are utilized as unary potential functions. 
The CPM used in~\cite{kong2020rotation} and~\cite{kong2019adaptive} is the version where $7\times 7$ convolutional kernels are replaced by three $3\times3$ convolutional kernels.
To make fair comparison, we follow their configurations and use their version of CPM as our preliminary pose estimator.
The fundamental difference between our method and~\cite{kong2020rotation} is that we have adopted our SIA-GCN instead of graphical models. As observed from Table~\ref{tab:state-of-the-art}, our method outperforms both~\cite{kong2020rotation, kong2019adaptive} on the Panoptic dataset. 
On the MHP dataset, our SiaPose also achieves the state-of-the-art level performance. The size of the MHP dataset is about five times the size of the Panoptic, making the MHP dataset an easier task and allows less room for improvement. Methods focused on modeling structural relationships between keypoints would benefit more from smaller and challenging datasets that require models to extrapolate beyond pose templates seen in the training data.

\textbf{Complexity analysis.} Regarding the size of the proposed models, the 5-head and 10-head models increase the model size by about 30\% and 40\%, respectively, compared to the 1-head model. The increment of the model size from 1-head to multiple heads is primarily due to the added pointer network, which is drawn in Fig.~\ref{fig:system_diagram}. However, going from 5-head to 10-head does not significantly increase model complexity. This is because the pointer network only needs to output 5 more scalers and the overall overhead mostly comes from adding more GCN layers, which are shallow and not associated with too many parameters (note that we use “channel-wise” 2D convolutions). 
It's also worth to point out that, using a 10-head SIA-GCN, our model
is about 80\% and 60\% the size of those in~\cite{kong2019adaptive} and~\cite{kong2020rotation}, respectively.

\textbf{Domain generalization of our model.} Table~\ref{tab:MPII} demonstrates the domain generalization ability of our model. All the models in Table~\ref{tab:MPII} are pretrained on Panoptic dataset, and then finetuned for about 40 epochs on the MPII+NZSL dataset. Consistent improvements over baselines are seen for all the ranges of PCK thresholds.

\begin{figure}[!h]
    \centering
    \includegraphics[width=0.9\textwidth]{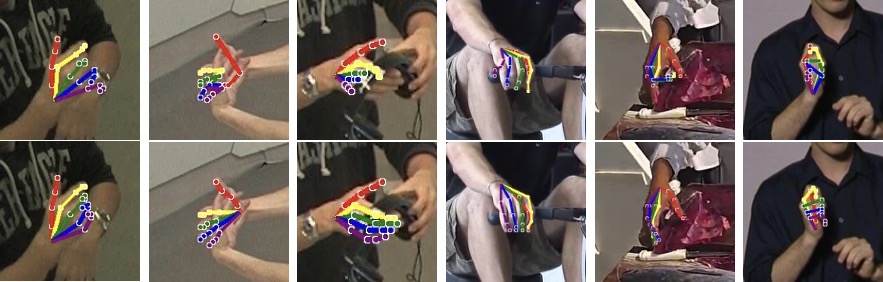}
    \caption{Qualitative results of baseline (top) and our model (bottom) on Panoptic and MPII.}
    \label{fig:qualitative}
\end{figure}

\begin{figure}[!h]
    \centering
    \includegraphics[width=0.9\textwidth]{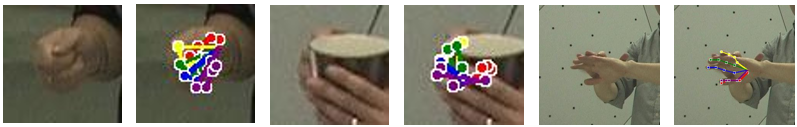}
    \caption{Failure cases of our model. Each pair contains an input image and its prediction.}
    \label{fig:failure_cases}
\end{figure}

\textbf{Qualitative results.} Some qualitative examples are given in Fig.~\ref{fig:qualitative}, which indeed shows that the SIA-GCN helps to enhance structural consistency and alleviate the spatial ambiguity. For example, in the third column, although the right hand is partially occluded by the earphone, our model could still correctly predict the position of all keypoints. We also show some failure cases of our model in  Fig.~\ref{fig:failure_cases}, which are due to very heavy occlusion and foreshortened view of a fist. 

\begin{table}[h]
\small 
  \centering
  \caption{Comparison to state-of-the-art methods.}
  \label{table:descrtotal} 
  \begin{threeparttable}
    \begin{tabular}{c|r|r|r|r|r|r|r}
    \hline
         PCK@ & 0.01& 0.02& 0.03& 0.04& 0.05& 0.06& mPCK \\
         \hline
         \hline
         \multicolumn{8}{c}{CMU Panoptic Hand Dataset}\\
         \hline
         R-MGMN~\cite{kong2020rotation} & 23.67 & 60.12 & 76.28 & 83.14 & 86.91 &89.47 &69.93 \\
         AGMN~\cite{kong2019adaptive} &23.90 &60.26& 76.21& 83.70& 87.72& 90.27 &70.34\\
         R-SiaPose (Ours) & 24.94 & 62.08& 77.83& 84.91& 88.78& 91.34& 71.65 \\
         \hline
         \hline
         \multicolumn{8}{c}{Large-scale Multiview 3D Hand Pose Dataset (MHP)}\\
         \hline
         R-MGMN~\cite{kong2020rotation} &41.51 &85.97 &93.71 &96.33 &97.51& 98.17& 85.53\\
         AGMN~\cite{kong2019adaptive} &41.38 &85.67 &93.96 &96.61 &97.77 &98.42  &85.63 \\
         R-SiaPose (Ours) & 41.27 & 85.89& 93.82& 96.43& 97.61&98.29& 85.56 \\
         \hline         
    \end{tabular}
  \end{threeparttable}
  \label{tab:state-of-the-art}
\end{table}

\begin{table}[h]
\small 
    \centering
    \caption{Domain generalization of our model to MPII+NZSL from Panoptic Dataset.}
    \begin{tabular}{c|r|r|r|r|r|r|r|r}
    \hline
    PCK@ & 0.01 & 0.02& 0.03&0.04& 0.05& 0.06& 0.07& 0.08 \\
    \hline
    \hline
         CPM & 8.05 & 23.78 & 37.74 & 48.00 & 55.65 & 61.68 & 66.58 &70.82 \\
         \hline

          R-SiaPose (Ours)  & 8.40 & 24.71 & 39.33 & 50.31 & 59.04 & 66.01 & 71.29&75.63 \\
          \hline
          Improvement &0.35& 0.93&1.59& 2.31& 3.39& 4.33& 4.71& 4.81 \\
         \hline
         \hline
         SHG & 11.72& 30.85& 44.82& 54.71 & 62.35& 68.48& 73.47& 77.61 \\
        \hline
         R-SiaPose (Ours) & 12.19& 33.34& 49.13& 59.86 & 67.83& 73.69& 78.26& 81.72 \\
        \hline
        Improvement &0.47& 2.49& 4.31& 5.15& 5.48& 5.21& 4.79& 4.11 \\
        \hline
        
    \end{tabular}
    \label{tab:MPII}
\end{table}{}

\section{Conclusion}
In this paper, we propose a novel spatial information aware graph neural network with 2D convolutions (SIA-GCN), which has the advantage of processing 2D spatial features for each node, with additional capability of learning different spatial relationships for different pair of neighboring nodes. We show the efficacy of our SIA-GCN in the 2D hand pose estimation task, by implementing a network which achieves the state-of-the-art performance. The SIA-GCN has the potential to generalise well to other tasks.

\clearpage
\bibliography{egbib}

\begin{thebibliography}{37}
\providecommand{\natexlab}[1]{#1}
\providecommand{\url}[1]{\texttt{#1}}
\expandafter\ifx\csname urlstyle\endcsname\relax
  \providecommand{\doi}[1]{doi: #1}\else
  \providecommand{\doi}{doi: \begingroup \urlstyle{rm}\Url}\fi

\bibitem[Baek et~al.(2018)Baek, In~Kim, and Kim]{baek2018augmented}
Seungryul Baek, Kwang In~Kim, and Tae-Kyun Kim.
\newblock Augmented skeleton space transfer for depth-based hand pose
  estimation.
\newblock In \emph{Proceedings of the IEEE Conference on Computer Vision and
  Pattern Recognition}, pages 8330--8339, 2018.

\bibitem[Cai et~al.(2018)Cai, Ge, Cai, and Yuan]{cai2018weakly}
Yujun Cai, Liuhao Ge, Jianfei Cai, and Junsong Yuan.
\newblock Weakly-supervised 3d hand pose estimation from monocular rgb images.
\newblock In \emph{Proceedings of the European Conference on Computer Vision
  (ECCV)}, pages 666--682, 2018.

\bibitem[Cai et~al.(2019)Cai, Ge, Liu, Cai, Cham, Yuan, and
  Thalmann]{cai2019exploiting}
Yujun Cai, Liuhao Ge, Jun Liu, Jianfei Cai, Tat-Jen Cham, Junsong Yuan, and
  Nadia~Magnenat Thalmann.
\newblock Exploiting spatial-temporal relationships for 3d pose estimation via
  graph convolutional networks.
\newblock In \emph{Proceedings of the IEEE International Conference on Computer
  Vision}, pages 2272--2281, 2019.

\bibitem[Chen and Yuille(2014)]{chen2014articulated}
Xianjie Chen and Alan~L Yuille.
\newblock Articulated pose estimation by a graphical model with image dependent
  pairwise relations.
\newblock In \emph{Advances in neural information processing systems}, pages
  1736--1744, 2014.

\bibitem[Chen et~al.(2020)Chen, Ma, Kong, Yan, Wu, Fan, and
  Xie]{chen2020nonparametric}
Yifei Chen, Haoyu Ma, Deying Kong, Xiangyi Yan, Jianbao Wu, Wei Fan, and
  Xiaohui Xie.
\newblock Nonparametric structure regularization machine for 2d hand pose
  estimation.
\newblock In \emph{The IEEE Winter Conference on Applications of Computer
  Vision}, pages 381--390, 2020.

\bibitem[Chen et~al.(2018)Chen, Wang, Peng, Zhang, Yu, and
  Sun]{chen2018cascaded}
Yilun Chen, Zhicheng Wang, Yuxiang Peng, Zhiqiang Zhang, Gang Yu, and Jian Sun.
\newblock Cascaded pyramid network for multi-person pose estimation.
\newblock In \emph{Proceedings of the IEEE conference on computer vision and
  pattern recognition}, pages 7103--7112, 2018.

\bibitem[Doosti(2019)]{DoostiSurvey}
Bardia Doosti.
\newblock Hand pose estimation: {A} survey.
\newblock \emph{CoRR}, abs/1903.01013, 2019.
\newblock URL \url{http://arxiv.org/abs/1903.01013}.

\bibitem[Doosti et~al.(2020)Doosti, Naha, Mirbagheri, and
  Crandall]{doosti2020hope}
Bardia Doosti, Shujon Naha, Majid Mirbagheri, and David Crandall.
\newblock Hope-net: A graph-based model for hand-object pose estimation.
\newblock \emph{arXiv preprint arXiv:2004.00060}, 2020.

\bibitem[Francisco Gomez-Donoso and Cazorla(2017)]{Francisco2017}
Sergio Orts-Escolano Francisco Gomez-Donoso and Miguel Cazorla.
\newblock Large-scale multiview 3d hand pose dataset.
\newblock \emph{ArXiv e-prints 1707.03742}, 2017.

\bibitem[Ge et~al.(2016)Ge, Liang, Yuan, and Thalmann]{ge2016robust}
Liuhao Ge, Hui Liang, Junsong Yuan, and Daniel Thalmann.
\newblock Robust 3d hand pose estimation in single depth images: from
  single-view cnn to multi-view cnns.
\newblock In \emph{Proceedings of the IEEE conference on computer vision and
  pattern recognition}, pages 3593--3601, 2016.

\bibitem[Ge et~al.(2018)Ge, Ren, and Yuan]{ge2018point}
Liuhao Ge, Zhou Ren, and Junsong Yuan.
\newblock Point-to-point regression pointnet for 3d hand pose estimation.
\newblock In \emph{Proceedings of the European Conference on Computer Vision
  (ECCV)}, pages 475--491, 2018.

\bibitem[Kipf and Welling(2016)]{kipf2016semi}
Thomas~N Kipf and Max Welling.
\newblock Semi-supervised classification with graph convolutional networks.
\newblock \emph{arXiv preprint arXiv:1609.02907}, 2016.

\bibitem[Kong et~al.(2019)Kong, Chen, Ma, Yan, and Xie]{kong2019adaptive}
Deying Kong, Yifei Chen, Haoyu Ma, Xiangyi Yan, and Xiaohui Xie.
\newblock Adaptive graphical model network for 2d handpose estimation.
\newblock In \emph{Proceedings of the British Machine Vision Conference
  ({BMVC})}, 2019.

\bibitem[Kong et~al.(2020)Kong, Ma, Chen, and Xie]{kong2020rotation}
Deying Kong, Haoyu Ma, Yifei Chen, and Xiaohui Xie.
\newblock Rotation-invariant mixed graphical model network for 2d hand pose
  estimation.
\newblock In \emph{The IEEE Winter Conference on Applications of Computer
  Vision}, pages 1546--1555, 2020.

\bibitem[Malik et~al.(2020)Malik, Abdelaziz, Elhayek, Shimada, Ali, Golyanik,
  Theobalt, and Stricker]{malik2020handvoxnet}
Jameel Malik, Ibrahim Abdelaziz, Ahmed Elhayek, Soshi Shimada, Sk~Aziz Ali,
  Vladislav Golyanik, Christian Theobalt, and Didier Stricker.
\newblock Handvoxnet: Deep voxel-based network for 3d hand shape and pose
  estimation from a single depth map.
\newblock \emph{arXiv preprint arXiv:2004.01588}, 2020.

\bibitem[Mueller et~al.(2017)Mueller, Mehta, Sotnychenko, Sridhar, Casas, and
  Theobalt]{mueller2017real}
Franziska Mueller, Dushyant Mehta, Oleksandr Sotnychenko, Srinath Sridhar, Dan
  Casas, and Christian Theobalt.
\newblock Real-time hand tracking under occlusion from an egocentric rgb-d
  sensor.
\newblock In \emph{Proceedings of the IEEE International Conference on Computer
  Vision}, pages 1284--1293, 2017.

\bibitem[Newell et~al.(2016)Newell, Yang, and Deng]{newell2016stacked}
Alejandro Newell, Kaiyu Yang, and Jia Deng.
\newblock Stacked hourglass networks for human pose estimation.
\newblock In \emph{European conference on computer vision}, pages 483--499.
  Springer, 2016.

\bibitem[Shuman et~al.(2013)Shuman, Narang, Frossard, Ortega, and
  Vandergheynst]{shuman2013emerging}
David~I Shuman, Sunil~K Narang, Pascal Frossard, Antonio Ortega, and Pierre
  Vandergheynst.
\newblock The emerging field of signal processing on graphs: Extending
  high-dimensional data analysis to networks and other irregular domains.
\newblock \emph{IEEE signal processing magazine}, 30\penalty0 (3):\penalty0
  83--98, 2013.

\bibitem[Simon et~al.(2017)Simon, Joo, Matthews, and Sheikh]{simon2017hand}
Tomas Simon, Hanbyul Joo, Iain Matthews, and Yaser Sheikh.
\newblock Hand keypoint detection in single images using multiview
  bootstrapping.
\newblock In \emph{Proceedings of the IEEE conference on Computer Vision and
  Pattern Recognition}, pages 1145--1153, 2017.

\bibitem[Song et~al.(2017)Song, Wang, Van~Gool, and Hilliges]{song2017thin}
Jie Song, Limin Wang, Luc Van~Gool, and Otmar Hilliges.
\newblock Thin-slicing network: A deep structured model for pose estimation in
  videos.
\newblock In \emph{Proceedings of the IEEE Conference on Computer Vision and
  Pattern Recognition}, pages 4220--4229, 2017.

\bibitem[Sun et~al.(2019)Sun, Xiao, Liu, and Wang]{sun2019deep}
Ke~Sun, Bin Xiao, Dong Liu, and Jingdong Wang.
\newblock Deep high-resolution representation learning for human pose
  estimation.
\newblock In \emph{Proceedings of the IEEE Conference on Computer Vision and
  Pattern Recognition}, pages 5693--5703, 2019.

\bibitem[Tompson et~al.(2014)Tompson, Jain, LeCun, and
  Bregler]{tompson2014joint}
Jonathan~J Tompson, Arjun Jain, Yann LeCun, and Christoph Bregler.
\newblock Joint training of a convolutional network and a graphical model for
  human pose estimation.
\newblock In \emph{Advances in neural information processing systems}, pages
  1799--1807, 2014.

\bibitem[Toshev and Szegedy(2014)]{toshev2014deeppose}
Alexander Toshev and Christian Szegedy.
\newblock Deeppose: Human pose estimation via deep neural networks.
\newblock In \emph{Proceedings of the IEEE conference on computer vision and
  pattern recognition}, pages 1653--1660, 2014.

\bibitem[Wan et~al.(2018)Wan, Probst, Van~Gool, and Yao]{wan2018dense}
Chengde Wan, Thomas Probst, Luc Van~Gool, and Angela Yao.
\newblock Dense 3d regression for hand pose estimation.
\newblock In \emph{Proceedings of the IEEE Conference on Computer Vision and
  Pattern Recognition}, pages 5147--5156, 2018.

\bibitem[Wan et~al.(2019)Wan, Probst, Gool, and Yao]{wan2019self}
Chengde Wan, Thomas Probst, Luc~Van Gool, and Angela Yao.
\newblock Self-supervised 3d hand pose estimation through training by fitting.
\newblock In \emph{Proceedings of the IEEE Conference on Computer Vision and
  Pattern Recognition}, pages 10853--10862, 2019.

\bibitem[Wang et~al.(2018)Wang, Peng, and Liu]{wang2018mask}
Yangang Wang, Cong Peng, and Yebin Liu.
\newblock Mask-pose cascaded cnn for 2d hand pose estimation from single color
  image.
\newblock \emph{IEEE Transactions on Circuits and Systems for Video
  Technology}, 29\penalty0 (11):\penalty0 3258--3268, 2018.

\bibitem[Wang et~al.(2019)Wang, Chen, Rathore, Shin, and
  Fowlkes]{wang2019geometric}
Zhe Wang, Liyan Chen, Shaurya Rathore, Daeyun Shin, and Charless Fowlkes.
\newblock Geometric pose affordance: 3d human pose with scene constraints.
\newblock \emph{arXiv preprint arXiv:1905.07718}, 2019.

\bibitem[Wang et~al.(2020)Wang, Shin, and Fowlkes]{wang2020predicting}
Zhe Wang, Daeyun Shin, and Charless~C Fowlkes.
\newblock Predicting camera viewpoint improves cross-dataset generalization for
  3d human pose estimation.
\newblock \emph{arXiv preprint arXiv:2004.03143}, 2020.

\bibitem[Wei et~al.(2016)Wei, Ramakrishna, Kanade, and
  Sheikh]{wei2016convolutional}
Shih-En Wei, Varun Ramakrishna, Takeo Kanade, and Yaser Sheikh.
\newblock Convolutional pose machines.
\newblock In \emph{Proceedings of the IEEE conference on Computer Vision and
  Pattern Recognition}, pages 4724--4732, 2016.

\bibitem[Wu et~al.(2020)Wu, Pan, Chen, Long, Zhang, and
  Philip]{wu2020comprehensive}
Zonghan Wu, Shirui Pan, Fengwen Chen, Guodong Long, Chengqi Zhang, and S~Yu
  Philip.
\newblock A comprehensive survey on graph neural networks.
\newblock \emph{IEEE Transactions on Neural Networks and Learning Systems},
  2020.

\bibitem[Xiao et~al.(2018)Xiao, Wu, and Wei]{xiao2018simple}
Bin Xiao, Haiping Wu, and Yichen Wei.
\newblock Simple baselines for human pose estimation and tracking.
\newblock In \emph{Proceedings of the European conference on computer vision
  (ECCV)}, pages 466--481, 2018.

\bibitem[Xiong et~al.(2019)Xiong, Zhang, Xiao, Cao, Yu, Zhou, and
  Yuan]{xiong2019a2j}
Fu~Xiong, Boshen Zhang, Yang Xiao, Zhiguo Cao, Taidong Yu, Joey~Tianyi Zhou,
  and Junsong Yuan.
\newblock A2j: Anchor-to-joint regression network for 3d articulated pose
  estimation from a single depth image.
\newblock In \emph{Proceedings of the IEEE International Conference on Computer
  Vision}, pages 793--802, 2019.

\bibitem[Yan et~al.(2018)Yan, Xiong, and Lin]{yan2018spatial}
Sijie Yan, Yuanjun Xiong, and Dahua Lin.
\newblock Spatial temporal graph convolutional networks for skeleton-based
  action recognition.
\newblock In \emph{Thirty-second AAAI conference on artificial intelligence},
  2018.

\bibitem[Yang et~al.(2016)Yang, Ouyang, Li, and Wang]{yang2016end}
Wei Yang, Wanli Ouyang, Hongsheng Li, and Xiaogang Wang.
\newblock End-to-end learning of deformable mixture of parts and deep
  convolutional neural networks for human pose estimation.
\newblock In \emph{Proceedings of the IEEE Conference on Computer Vision and
  Pattern Recognition}, pages 3073--3082, 2016.

\bibitem[Yuan et~al.(2018)Yuan, Garcia-Hernando, Stenger, Moon, Yong~Chang,
  Mu~Lee, Molchanov, Kautz, Honari, Ge, et~al.]{yuan2018depth}
Shanxin Yuan, Guillermo Garcia-Hernando, Bj{\"o}rn Stenger, Gyeongsik Moon,
  Ju~Yong~Chang, Kyoung Mu~Lee, Pavlo Molchanov, Jan Kautz, Sina Honari, Liuhao
  Ge, et~al.
\newblock Depth-based 3d hand pose estimation: From current achievements to
  future goals.
\newblock In \emph{Proceedings of the IEEE Conference on Computer Vision and
  Pattern Recognition}, pages 2636--2645, 2018.

\bibitem[Zhao et~al.(2019)Zhao, Peng, Tian, Kapadia, and
  Metaxas]{zhao2019semantic}
Long Zhao, Xi~Peng, Yu~Tian, Mubbasir Kapadia, and Dimitris~N Metaxas.
\newblock Semantic graph convolutional networks for 3d human pose regression.
\newblock In \emph{Proceedings of the IEEE Conference on Computer Vision and
  Pattern Recognition}, pages 3425--3435, 2019.

\bibitem[Zimmermann and Brox(2017)]{zimmermann2017learning}
Christian Zimmermann and Thomas Brox.
\newblock Learning to estimate 3d hand pose from single rgb images.
\newblock In \emph{Proceedings of the IEEE International Conference on Computer
  Vision}, pages 4903--4911, 2017.

\end{thebibliography}
\end{document}